\documentclass[letterpaper]{article} 
\usepackage[preprint]{aaai2027}       

\usepackage[hyphens]{url}  
\usepackage{graphicx} 
\usepackage{amsmath}
\usepackage{amssymb}
\usepackage{amsthm}
\usepackage{natbib}  
\usepackage{caption} 
\usepackage{algorithm}
\usepackage{algorithmic}
\usepackage{listings}
\usepackage{xcolor}
\theoremstyle{definition}
\newtheorem{definition}{Definition}
\theoremstyle{plain}

\usepackage{newfloat}
\usepackage{listings}
\DeclareCaptionStyle{ruled}{labelfont=normalfont,labelsep=colon,strut=off} 
\floatstyle{ruled}
\newfloat{listing}{tb}{lst}{}
\floatname{listing}{Listing}

  \lstdefinelanguage{PDDL}{
    morekeywords={
      define,domain,problem,requirements,types,predicates,action,parameters,
      precondition,effect,and,or,not,when,forall,exists,objects,init,goal
    },
    sensitive=false,
    morecomment=[l]{;},
    morestring=[b]"
  }

\usepackage{xcolor}
\usepackage{minted}

\definecolor{pddlbg}{HTML}{F7F8FA}
\definecolor{pddlframe}{HTML}{D0D7DE}

\newminted[pddl]{pddl}{
  fontsize=\scriptsize,
  bgcolor=pddlbg,
  frame=single,
  rulecolor=pddlframe,
  framesep=2mm,
  breaklines=true,
  breakanywhere=false,
  autogobble=true,
  tabsize=2,
  samepage=false
}

\usepackage{booktabs}

\providecommand{\suppref}[1]{the supplementary material}
\providecommand{\Suppref}[1]{The supplementary material}
\providecommand{\mainfigref}[1]{the corresponding figure in the main paper}

\renewcommand{\suppref}[1]{Appendix~\ref{#1}}
\renewcommand{\Suppref}[1]{Appendix~\ref{#1}}
\renewcommand{\mainfigref}[1]{Fig.~\ref{#1}}

\title{When Automata Meet Streams: Temporal Logic Compilation for \\ Stream-Based Robotics Task and Motion Planning}
\author{
    Sayem Nazmuz Zaman, Cyrus Neary
}
\affiliations{
    Department of Electrical and Computer Engineering \\
    University of British Columbia\\

    Vancouver, BC, Canada\\
}

\begin{document}
\maketitle

\begin{abstract}
\label{sec:abs}
Stream-based robotics Task and Motion Planning (TAMP) integrates discrete symbolic planning with dynamically generated continuous geometric parameters, such as poses, grasps, and trajectories.
However, stream-based planners typically reason only about goal reachability, whereas long-horizon tasks also demand adherence to temporal specifications, such as safety-critical ordering, invariance, and liveness constraints.
No methods currently exist to enforce such temporal constraints for stream-based solvers because streams generate an expanding geometric object set via iterative stream refinement loops during planning, rendering existing temporal-logic compilation techniques incompatible.
We therefore present Synchronous Action Monitoring with Token Destruction (SAM-TD), a compilation method that enforces arbitrary Linear Temporal Logic over finite traces ($\textrm{LTL}_f$) specifications in stream-based TAMP.
SAM-TD translates arbitrary $\textrm{LTL}_f$ constraints into automata and embeds regressed automaton guards into action schemas, which are pre-specified before planning begins.
By doing so, SAM-TD can handle objects generated by streams during planning, thus circumventing the need to enumerate a fixed object set or modify the underlying planner.
During search, SAM-TD synchronously updates automaton states and uses a validity token shared across all automata to prune constraint-violating branches. 
We show that SAM-TD supports dynamically generated stream objects from iterative stream refinements during plan search.
Experimental results provide the first ever demonstration of stream-based TAMP under $\textrm{LTL}_f$ constraints in three robotics PDDLStream environments.
Furthermore, on standard discrete PDDL benchmarks, SAM-TD is competitive with state-of-the-art temporal-constraint compilation methods.
\end{abstract}

\section{Introduction}
\label{sec:intro}
\begin{figure*}[t]
\centering
\includegraphics[width=\textwidth]{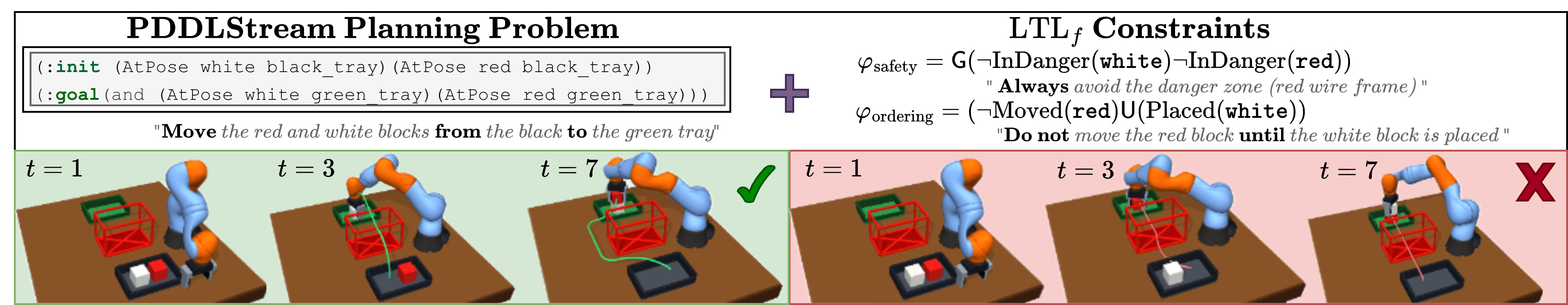}
\caption{
Illustrative example with two LTL$_f$ constraints: $\varphi_{\text{safety}}$ keeps the white and red blocks out of the marked danger region (red wireframe), and $\varphi_{\text{ordering}}$ forbids moving the red block until the white block is placed.
The planning problem requires the robot to move both of the blocks from the black tray to the green tray on the other side of the table.
The unconstrained plan (red) is symbolically valid but carries the blocks straight through the danger region (red trace) and moves them in the incorrect sequence. 
With SAM-TD compiling both constraints into the PDDLStream task, an unmodified planner returns a plan (green) whose sampled trajectory routes around the region (green trace) while respecting the ordering.
}
\label{fig:frame-motion}
\end{figure*}
Robots deployed in homes, laboratories, and factories are expected to complete long-horizon tasks while satisfying temporal requirements which are often safety-critical.
A chemistry robot \cite{Yoshikawa2022Chemistry, Darvish2024Organa}, for example, might have to turn on the fume cupboard before carrying out hazardous chemical reactions inside, a surveillance drone has to always ensure that it can return to a charging station, and a Mars rover needs to eventually send data it collects to a central station during its task execution. 
A well-studied formal language for representing such temporal specifications is Linear Temporal Logic over finite traces (LTL$_f$) \cite{Pnueli1977, DeGiacomo2013}.

Solving such long-horizon robotics tasks also requires jointly reasoning about discrete, high-level actions and continuous, low-level geometric quantities: the central concern of Task and Motion Planning (TAMP).
Classical task planners commonly represent the discrete component using the Planning Domain Definition Language (PDDL), where a planning problem is defined over a fixed and finite set of objects. 
PDDLStream \cite{Garrett2020} extends PDDL-based planning to continuous domains using black-box generators, called \emph{streams}, that produce geometric objects (such as grasps, placements, and trajectories) on-demand during plan search.

In classical PDDL domains, LTL$_f$ specifications can be compiled into planning constructs \cite{Baier2006,Torres2015,Camachoetal2017}, allowing them to be enforced by unmodified planners.
These compilation techniques, however, do not extend directly to stream-based TAMP problems (PDDLStream) because they assume the planning problem has a fixed and finite object set.
By contrast, streams dynamically generate and iteratively refine previously unknown geometric objects during planning, resulting in an expanding (and potentially infinite) object set.
This creates a fundamental incompatibility between existing temporal constraint compilation techniques and stream-based TAMP.
Pre-sampling a finite library of geometric objects does not provide an alternative, since the required objects may not be known in advance and are often plan dependent.
Additionally, generating a sufficiently large library can be prohibitively expensive for both sampling and plan search.
Consequently, there are no methods for enforcing arbitrary LTL$_f$ constraints in stream-based TAMP.

In this work, we introduce Synchronous Action Monitoring with Token Destruction (SAM-TD): a compilation framework that embeds LTL$_f$ constraint violation monitors directly into the high-level action specifications (lifted action schemas) of a stream-based TAMP problem.
By compiling the monitoring logic into lifted action schemas, which are pre-specified before planning begins, SAM-TD can handle objects generated dynamically by streams during planning, thus circumventing the need to enumerate a fixed object universe or modify the underlying planner.

SAM-TD translates every LTL$_f$ constraint into a deterministic finite automaton (DFA).
Since the automaton guards describe the state produced \emph{after} an action, SAM-TD regresses each guard through the action's effects to obtain an equivalent condition on the state and the action parameters \textit{before} the action's execution.
Conditional effects added to the action then update the corresponding automaton state synchronously with the original action.
To monitor violations, SAM-TD introduces a validity token, $\mathit{Valid}$, carried by every feasible plan.
The token is initially present, required as part of the goal, and can never be added by any action.
If an action drives any DFA into a sink state, the action's conditional effect permanently deletes $\mathit{Valid}$ (Token Destruction).
The violating action therefore remains applicable, but executing it makes the goal unreachable, allowing standard delete-relaxation heuristics to prune the resulting dead end. Constraints that have not been violated must additionally terminate in accepting automaton states.

We summarize our contributions as follows: 
(1) We introduce the first compilation technique to embed arbitrary LTL$_f$ constraints into stream-based TAMP problems, guaranteeing that any plan returned by a stream-based TAMP solver satisfies all specified LTL$_f$ constraints.  
(2) We show that the Token Destruction mechanism can be viewed as a two-state monitor automaton over DFA sink transitions.
(3) We provide the first empirical evaluation of LTL$_f$-constrained stream-based TAMP problems and identify key bottlenecks in doing so.
(4) We show that our stream-based TAMP compilation technique is backward-compatible with, and competitive on classical PDDL benchmarks. 

\section{Background and Related Works}
\label{sec:background}
\subsection{LTL$_f$ and Deterministic Finite Automata}
\label{subsec:background-ltlf}
LTL$_f$ \cite{DeGiacomo2013}, the finite-trace version of LTL \cite{Pnueli1977}, extends propositional logic over a set of atoms $AP$ with the operators \emph{next} ($\mathsf{X}$) and \emph{until} ($\mathsf{U}$): 
\begin{align*}
    \varphi ::= p \mid \neg\varphi \mid \varphi_1 \wedge \varphi_2 \mid \mathsf{X}\varphi \mid \varphi_1\,\mathsf{U}\,\varphi_2, \qquad p \in AP,
\end{align*}
with \emph{eventually} $\mathsf{F}\varphi \equiv \mathit{true}\,\mathsf{U}\,\varphi$, \emph{always} $\mathsf{G}\varphi \equiv \neg\mathsf{F}\neg\varphi$, \emph{release} $\mathsf{R}$, and \emph{weak next} $\mathsf{WX}$.
The truth values for the formulas are evaluated over finite traces $\sigma = s_0 \cdots s_n$, with $n \geq 0$.
Every LTL$_f$ formula $\varphi$ can be translated into a Deterministic Finite Automaton (DFA)
\begin{align*}
    \mathcal{D}_\varphi = \langle Q, 2^{AP}, q_0, \delta, F \rangle
\end{align*}
that accepts exactly the traces satisfying $\varphi$ \cite{DeGiacomo2013}. 
Here, $Q$ is a finite set of states, $AP$ is the set of atomic propositions, $2^{AP}$ is the input alphabet of their possible truth assignments, $q_0 \in Q$ is the initial state, $\delta \colon Q \times 2^{AP} \rightarrow Q$ is the transition function, and $F \subseteq Q$ is the set of accepting states.

The DFA acts as a \emph{monitor} for $\varphi$, where reading the trace in sequence transitions the automaton through its states, and the trace satisfies $\varphi$ if and only if its run ends in an accepting state. 
A trace that transitions the DFA into a sink state, a non-accepting state from which no accepting state is reachable, violates $\varphi$. 
The LTL$_f$ to DFA construction is doubly exponential in $|\varphi|$ in the worst case \cite{DeGiacomo2013}.

In this work, a temporal specification is a finite constraint set, $C = \{\varphi_1, \dots, \varphi_m \}$ of LTL$_f$ formulas. 
We abbreviate the DFA of $\varphi_j$ as $\mathcal{D}_j$. 
A plan $\pi$ (a sequence of symbolic actions) satisfies $C$ if and only if every $\mathcal{D}_j$ accepts the trace $\tau(\pi)$ that $\pi$ induces.

\subsection{Classical Planning}
\label{subsec:classical-TAMP}
A lifted classical planning task is a tuple $\langle \mathbf{P}, \mathbf{O}, \mathcal{A}, I, G \rangle$ with predicate symbols $\mathbf{P}$, object set $\mathbf{O}$, action schemas $\mathcal{A}$, initial state $I$, and goal condition $G$
\cite{McDermott1998}. 
Substituting objects for the parameters of an atom or schema is called \textbf{grounding}; an atom or action with no remaining parameters is \textbf{ground}.
$\text{Ground}(\mathbf{P}, \mathbf{O})$ denotes the set of all ground atoms, and a state $s \subseteq \text{Ground}(\mathbf{P}, \mathbf{O})$ is the set of ground atoms currently true. 
 Each action schema $a = \langle \mathrm{params}(a), \mathrm{pre}(a), \mathrm{eff}^{+}(a), \mathrm{eff}^{-}(a) \rangle$ has a precondition and add/delete effects over its parameters and may carry conditional effects $(c \rhd e)$, which apply $e$ only when $c$ holds in the state where the action executes. 
Executing an applicable ground action $a$ in state $s$ yields the successor $\gamma(s, a)$. 
A plan $\pi = a_1, \ldots, a_n$ is applicable in $I$ if it induces the \emph{state sequence} (trace) $\tau(\pi) = s_0 s_1 \cdots s_n$, with $s_0 = I$ and $s_i = \gamma(s_{i-1}, a_i)$; 
$\pi$ solves the task iff $s_n \models G$. 
Reading each $s_i$ as the truth assignment it induces over $AP$ turns this state sequence into the trace $\tau(\pi)$, the object over which our temporal constraints are evaluated.

\subsection{Action Regression via Weakest Preconditions}
\label{subsec:WP}
Action Regression is the predicate-transformer \cite{Dijkstra1976} view of action execution: 
given an action $a$ and a post-condition $\psi$, which is a propositional formula over ground atoms, the weakest precondition regression $\mathrm{Reg}_a(\psi)$ gives us the least restrictive condition on the action's pre-state that guarantees $\psi$ holds after executing $a$ \cite{Rintanen2008}. 
Regression is classically used for backward search and invariant synthesis \cite{Rintanen2008}, and more recently for trajectory-constraint compilation \cite{Bonassi2021On, Mantenoglou2026}.
We use it for the same purpose in the stream-based setting (\S~\ref{subsec:Lifted-Action-Regression}):
the transition conditions of the DFA refer to the state \emph{after} an action, and regression turns each of them into an equivalent check on the state \emph{before} the action, so that the planner can decide whether to apply the action.

\subsection{Stream-based Task and Motion Planning}
\label{subsec:stream-TAMP}
Stream-based TAMP \cite{Garrett2020, Garrett2021} extends classical planning tasks with continuous geometric reasoning over robot configurations, poses, grasps, and trajectories using \emph{streams}: declaratively specified black-box generators $s = \langle \mathrm{inp}(s), \mathrm{dom}(s), \mathrm{out}(s), \mathrm{cert}(s) \rangle$ that consume objects satisfying the domain facts $\mathrm{dom}(s)$ (e.g., a pose and a grasp) and dynamically generate new geometric objects (e.g., an inverse-kinematics solution) together with \textbf{stream-certified facts} $\mathrm{cert}(s)$ declaring what the outputs satisfy (e.g., $\mathit{Kin}(b, q, p, g)$). 
Algorithms such as PDDLStream's \emph{Adaptive} algorithm  \cite{Garrett2020} alternate between (i) an \emph{optimistic search} phase, which plans over a finite PDDL abstraction containing placeholder objects for not-yet-sampled stream outputs, and (ii) a \emph{plan refinement} phase, which calls samplers along candidate plan skeletons to bind the placeholders, feeding failures back into search.
As we shall see later on, this plan--sample--refine loop creates three requirements for any stream-supportive temporal-constraint compilation method.

Although classical TAMP solvers can compile the lifted PDDL task into a grounded representation before search begins (\textit{eager} grounding), stream-based TAMP solvers cannot: the object set grows during planning as streams produce values. 
Stream-based solvers therefore replace the classical planner's translator with their own instantiation layer, which grounds an action only once its precondition predicates are ground and its geometric parameters are either pre-specified or streamed.
As a consequence of this stream-specific grounding requirement and the plan--sample--refine loop inherent to most stream-based solvers, any stream-supportive temporal constraint compilation method must satisfy all the following requirements:

\begin{itemize}
    \item 
    \textbf{Schema-level, stream-compatible encoding: } The compilation must operate on lifted action schemas rather than ground operators, since the object set is unknown before planning. Any schema modification must remain sound as streams introduce new objects, and must respect the instantiation layer's semantics (parameter–constant matching, unbound action parameters, no equality checks; see supplementary material).
    \item 
    \textbf{Preservation of the domain-action skeleton: } One original domain action must correspond to one monitored trace step and no synchronization or bookkeeping actions may be inserted as these can break the plan--sample--refine loop.
    \item 
    \textbf{Sampling-aware pruning: } Expanding a symbolic action during search may trigger costly geometric sampling. A useful encoding must therefore let the planner avoid searching constraint-violating branches.    
\end{itemize}
SAM-TD (\S~\ref{sec:method}) satisfies all these requirements by construction.
Further discussion is in the supplementary material.
\begin{figure*}[t]
\centering
\includegraphics[width=\textwidth]{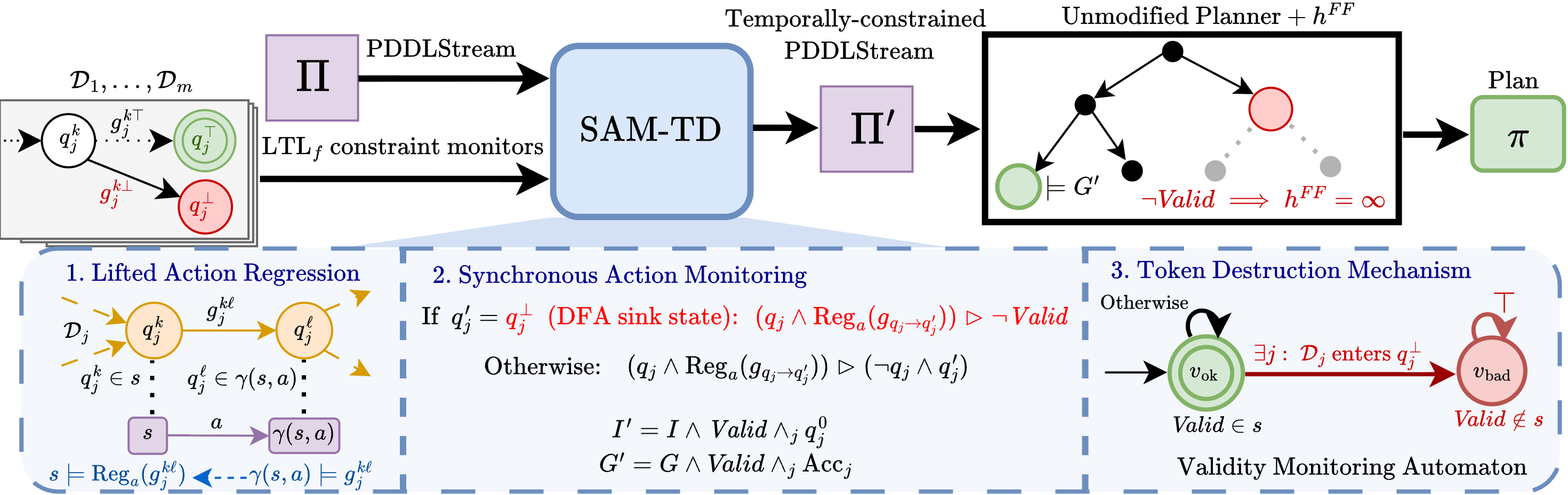} 
\caption{SAM-TD pipeline. \textbf{Left: } 
Inputs are an unmodified PDDLStream problem $\Pi$ and DFA monitors for each LTL$_f$ constraint. 
\textbf{Inside SAM-TD: } (1) Each transition guard $g^{k \ell}_j$ constrains the post-action state, $\gamma(s, a) \vDash g^{k \ell}_j$, and is regressed through every lifted action schema into the equivalent pre-action condition: $s \vDash \mathrm{Reg}_a(g^{k \ell}_j)$ (\S~\ref{subsec:Lifted-Action-Regression}).
(2) The regressed transitions are added to action schemas as conditional effects that advance the monitor in lockstep with the action (\S~\ref{subsec:SAM-TD}) during plan search.
(3) Transitions into a sink delete a $\textit{Valid}$ token: the compiled state of a two-state violation monitor tracking all DFAs at once (\S~\ref{subsec:VMA}). 
The compiled goal requires the $\textit{Valid}$ token and all DFAs in accepting states. 
\textbf{Right: } The compiled problem $\Pi'$ is solved by an unmodified planner with $h^{\text{FF}}$. Since no action re-adds $\textit{Valid}$, any state that loses it sets $h^{\text{FF}} = \infty$, so violating branches (red) are pruned before geometric sampling is spent on them.
The returned plan $\pi$ satisfies all constraints.
}
\label{fig:samtd-pipeline}
\end{figure*}
\subsection{Related Works}
\label{subsec:related-work}
Classical temporal-goal compilations commonly augment a planning task with automaton state. 
\citet{Baier2006} compile LTL$_f$ goals into Non-deterministic Finite Automata (NFA), and update the reachable NFA states via action conditional effects defined over ground atoms. 
\citet{Edelkamp2006Compilation}, \citet{Torres2015}, and \citet{Camachoetal2017} each track different types of automatons through dedicated \textit{synchronization-update} actions interleaved between domain actions, breaking PDDLStream's plan--sample--refine loop. 
\citet{DeGiacomoetal2022} compile Pure-Past LTL (ppLTL) by tracking selected past-subformula
values through derived predicates and effects 
which are defined over ground atoms as in
\citet{Baier2006}. 
TCORE \cite{Bonassi2021On} compiles PDDL3 trajectory constraints via regression, but only over fully grounded tasks. 
Its lifted successors LiftedTCORE and LCC \cite{Mantenoglou2026} avoid grounding, but LiftedTCORE regression matches action parameters to constraint objects through equality checks, breaking stream-specific semantics while LCC adds bookkeeping actions.
In summary, existing temporal constraint compilations generate artifacts that make them incompatible for stream-based TAMP.
Addressing this incompatibility is the primary focus of our work.
\section{Methodology}
\label{sec:method}
\subsection{Problem Formulation}
\label{subsec:problem}
We build on the PDDLStream formalism \cite{Garrett2020}. 
A \textbf{PDDLStream problem} is a tuple $\Pi = \langle \textbf{P}, \textbf{O}, \mathcal{A}, I, G, \mathcal{S} \rangle$, with the same formulation as a classical planning problem (\S~\ref{subsec:classical-TAMP}) extended with a set of streams, $\mathcal{S}$ (\S~\ref{subsec:stream-TAMP}).
Streams recursively extend $I$ with certified facts over generated objects, yielding the (possibly infinite) optimistic closure $I^{*}$.
A \textbf{binding} $\theta$ maps every parameter of an action schema $a \in \mathcal{A}$ to an object and thus we use $a(\theta)$ to denote the resulting ground action.
A \textbf{solution} is a finite ground action sequence $\pi = \langle a_1(\theta_1), \ldots, a_n(\theta_n) \rangle$ whose pre-image is contained in $I^{*}$ and whose final state satisfies $G$. 
We write $s_0 \xrightarrow{a_1} s_1 \xrightarrow{a_2} \cdots \xrightarrow{a_n} s_n$ for the induced state sequence, with $s_0 = I$.
Temporal constraints are stated over the discrete part of the state (\emph{task-level}) and never about the stream-certified facts (\emph{motion-level}).
Each state $s_i$ induces a propositional valuation $L(s_i) = \{ p \in \mathit{AP} \mid p \in s_i \}$ which is the set of monitored atoms that are true in $s_i$.
If $\pi$ induces the state sequence $s_0, \dots, s_n$, its temporal trace is
\begin{align*}
 \tau(\pi)=L(s_0)L(s_1)\cdots L(s_n).
\end{align*}

\begin{definition}[LTL$_f$-constrained PDDLStream problem]
\label{def-1}
An LTL$_f$-constrained PDDLStream problem is a pair $(\Pi,C)$,
where $\Pi$ is a PDDLStream problem and $C=\{\varphi_1,\ldots, \varphi_m\}$ is a finite set of $m$ LTL$_f$ formulas over
$\mathit{AP}$.
Let
\begin{align*}
    \pi = \langle a_1(\theta_1),\ldots,a_n(\theta_n)\rangle
\end{align*}
be a finite sequence of ground actions, and let
\begin{align*}
s_0
\xrightarrow{a_1(\theta_1)}
s_1
\xrightarrow{a_2(\theta_2)}
\cdots
\xrightarrow{a_n(\theta_n)}
s_n
\end{align*}
be the state sequence induced by executing $\pi$ in $\Pi$.
The \emph{task-level trace induced by $\pi$} is the finite word
\begin{align*}
\tau_{\Pi}(\pi)
=
L(s_0)L(s_1)\cdots L(s_n)
\in (2^{\mathit{AP}})^{n+1}.
\end{align*} 
Then the plan $\pi$ is a solution of $(\Pi,C)$ if and only if
\begin{enumerate}
    \item $\pi$ is a solution of $\Pi$; and
    \item $\tau_{\Pi}(\pi)\models \varphi_j$ for every
    $j\in\{1,\ldots,m\}$.
\end{enumerate}
\end{definition}
Our objective, therefore, is a \emph{compilation}:  $(\Pi, C) \mapsto \Pi'$ whose solutions are exactly the solutions of $(\Pi,C)$ and whose execution traces preserve the one-domain-action-to-one-trace-transition correspondence above.
In particular, the compilation must not insert trace-visible monitoring
actions between consecutive domain actions, since doing so would change
the semantics of step-sensitive LTL$_f$ formulas such as $\mathsf{X}$ (\textit{next}) and also break the plan-sample-refine loop.

Additionally, because compilation alters \textit{only} the PDDL domain and problem, any PDDLStream planning algorithm (e.g., Incremental, Focused, Binding, Adaptive) can be used to solve the compiled problem \(\Pi'\).

Fig.~\ref{fig:samtd-pipeline} (Inside SAM-TD) summarizes the three main ideas of our compilation. SAM-TD regresses successor state guards, updates every monitor inside the original domain action, and makes a constraint violation permanently visible to the planner by deleting $\mathit{Valid}$. 
\subsection{From LTL$_f$ Constraints to Automata}\label{subsec:LTLf-to-DFA}
Following \S~\ref{subsec:background-ltlf}, we translate each LTL$_f$ constraint $\varphi_j$ into its DFA, $\mathcal{D}_j = \langle Q_j, 2^{{AP}_j}, q^0_j, \delta_j, F_j \rangle$ using LTLf2DFA and MONA \cite{Henriksenetal1995, fuggitti-ltlf2dfa}. 
We write $q^k_j$ for states of $\mathcal{D}_j$, with $q^0_j$ as the initial state. 
MONA returns each automaton in symbolic form where DFA transitions are grouped by guarded edges.
For each pair $q_j^k, q_j^{\ell} \in Q_j$, the guard $g_j^{k \ell}$ is a Boolean formula over ${AP}_j$ such that
\begin{align*}
    \delta_j(q^k_j,\sigma)=q_j^{\ell}
    \quad\Longleftrightarrow\quad
    \sigma \models g^{k \ell}_j
\end{align*}
where $\delta_j$ is the DFA transition function and \(\sigma \in 2^{AP_{j}}\) is a truth assignment to \(AP_{j}\), serving as the input symbol that either satisfies or falsifies the guard formula. 
The guarded transition set is $\mathcal{T}_j =
\{(q_j^k,g_j^{k \ell},q_j^{\ell}) \mid q_j^k ,q_j^{\ell} \in Q_j,\; g_j^{k\ell} \not\equiv \bot \}.$
Because $\mathcal{D}_j$ is deterministic and total, the guards on edges leaving any fixed state $q$ are mutually exclusive and jointly exhaustive; they partition the alphabet $2^{\mathit{AP}_j}$, so at each time-step exactly one outgoing edge fires. 
If some finite prefix of a trace has no continuation that can satisfy $\varphi_j$, the DFA collapses all such prefixes into a single non-accepting state from which no accepting state is reachable. 
We write $q^{\bot}_j$ for this rejecting sink state; a run that enters $q^{\bot}_j$ has irrecoverably violated the constraint.
Each transition $(q_j^k, g^{k \ell}_{j}, q_j^{\ell}) \in \mathcal{T}_j$ is classified by its target:
\begin{align*}
    \mathrm{type}(q_j^k, g^{k \ell}_{j}, q_j^{\ell}) = \begin{cases} \textbf{self} & q_j^k = q_j^{\ell} \\ 
    \textbf{sink} & q_j^{\ell} = q^{\bot}_j,\\
    \textbf{forward} & \text{otherwise.} \end{cases}
\end{align*}
Before any domain action executes, the monitor must observe the initial planning state. 
We therefore set  $q_j^{\mathrm{start}} =\delta_j(q_j^0,L(s_0))$ during compilation, since by Definition~\ref{def-1}, the trace begins with $L(s_0)$ and the automaton must read this first letter before it begins monitoring actions. 
If $q^{\mathrm{start}}_j = q^{\bot}_j$, the initial state alone already violates $\varphi_j$ and the compiled problem must be unsolvable.
Our compilation reproduces exactly this semantics: $\mathit{Valid}$
is then initially false and can never be added, so the compiled goal is unreachable (\S~\ref{subsec:SAM-TD}).
\subsection{Lifted Action Regression}\label{subsec:Lifted-Action-Regression}
A DFA guard is evaluated in the successor state; however, the planner must decide \emph{before} executing an action which transition that action will cause. 
We accordingly translate every guard into an equivalent condition on the pre-action state, separately for each action schema.
For a \textit{ground} action, regression is well defined \cite{Rintanen2008}: Let $a(\theta)$ be an applicable ground action and $g$ be some DFA guard over ${AP}_j$ for DFA $\mathcal{D}_j$:
\begin{align*}
    s \models \mathrm{Reg}_{a(\theta)}(g) \quad\Longleftrightarrow\quad \gamma(s,a(\theta)) \models g
\end{align*}
$\mathrm{Reg}_{a(\theta)}(g)$ is a Boolean formula, and its atoms are drawn from ${AP}_j$.
Recall, however, that our compilation must operate on lifted action schemas to allow streams to introduce geometric objects during planning.  
The main complication is that a DFA guard can mention a \textit{named object} (e.g., $\mathit{moved}(\mathit{blue})$), whereas the lifted action effect mentions a \textit{parameter} (e.g., $\mathit{moved}(?b)$).
So lifted regression must reason about an object's \emph{identity} rather than the object itself. 
For every named object $o$ mentioned in a DFA guard, we introduce a \textbf{static identity predicate} $\mathsf{Is}\text{-}o(\cdot)$,
with the single initial fact $\mathsf{Is}\text{-}o(o)$. 
With the identity atoms in place, $\mathrm{Reg}_a(g)$ is a Boolean formula over the guard's atoms in $AP$ together with identity atoms over $\mathrm{params}(a)$, computed once per schema.
We formally write lifted action regression as follows:
\begin{align*} 
s \models \mathrm{Reg}_a(g)[\theta] \quad\Longleftrightarrow\quad
\gamma(s,a(\theta)) \models g
\end{align*}
where $\mathrm{Reg}_a(\cdot)[\theta]$ replaces each parameter by its object under the binding $\theta$.
\paragraph{Regression of a ground atom}
Let $a$ be an action schema and let $p(\bar{o}) = p(o_1,\ldots,o_k)$ be a ground guard atom. For an action argument $t$ and object $o$, define
\begin{align*}
    \mathrm{Eq}(t,o)=
    \begin{cases}
    \mathsf{Is}\text{-}o(t) & \text{if } t \text{ is a parameter},\\
    \top & \text{if } t=o,\\
    \bot & \text{otherwise.}
\end{cases}
\end{align*}
The add- and delete-match conditions for $p(\bar{o})$ are
\begin{align*}
\mathrm{Add}_a^{p(\bar{o})}
&=
\bigvee_{p(\bar{t}) \in \mathit{eff}^{+}(a)}
\bigwedge_{\ell=1}^{k} \mathrm{Eq}(t_\ell,o_\ell),\\
\mathrm{Del}_a^{p(\bar{o})}
&=
\bigvee_{p(\bar{t}) \in \mathit{eff}^{-}(a)}
\bigwedge_{\ell=1}^{k} \mathrm{Eq}(t_\ell,o_\ell).
\end{align*}
Then the regression of $p(\bar{o})$ through $a$ is
\begin{align*}
    \mathrm{Reg}_a(p(\bar{o})) = 
    \mathrm{Add}_a^{p(\bar{o})}
    \vee
    \left(
    p(\bar{o}) \wedge \neg \mathrm{Del}_a^{p(\bar{o})}
    \right)
\end{align*}
We present the unconditional case; effects that are themselves conditional regress analogously \cite{Rintanen2008}.
Intuitively, $\mathrm{Reg}_a(p(\bar{o}))$ states that the atom holds
after $a$ iff $a$ makes it true, or it already held and $a$ does not
delete it.
Regression is applied compositionally: a compound guard is regressed by regressing its atoms and reassembling the results with the same Boolean structure \cite{Rintanen2008}.
\subsection{SAM-TD Compilation}\label{subsec:SAM-TD}
\begin{figure*}[t]
\centering
\includegraphics[width=\textwidth]{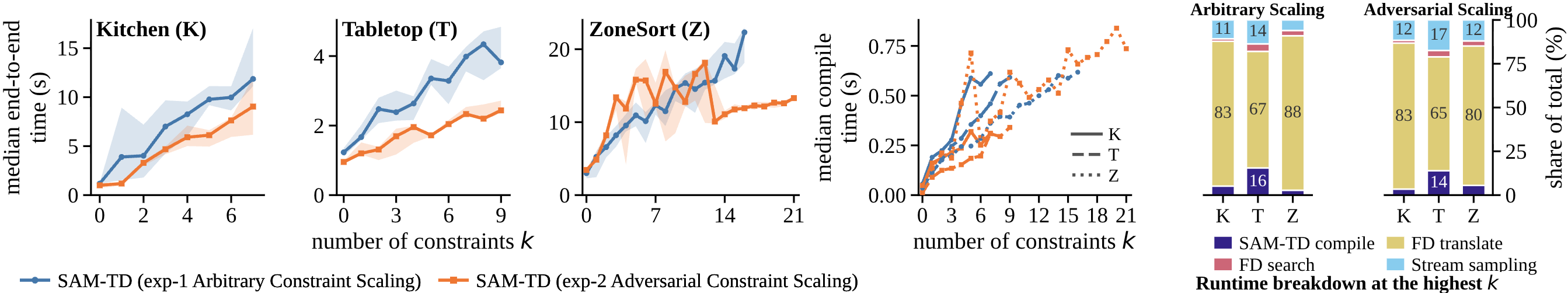}
\caption{SAM-TD median end-to-end (left) and median compilation (center) time in stream-based TAMP under adversarial and arbitrary constraint scaling (Experiment 1 and Experiment 2). 
Runtime breakdown (right) shows per-component medians as stacked areas. 
Fast Downward translation, not the SAM-TD compilation itself, dominates cost of end-to-end compilation and planning.}
\label{fig:pddlstream-scaling}
\end{figure*}
\begin{algorithm}[t]
\caption{\textsc{SAM-TD}($\Pi,C$)}
\label{alg:gr-compile}
\small
\begin{algorithmic}[1]
\REQUIRE $\Pi=\langle\mathbf{P},\mathbf{O},\mathcal{A},I,G,\mathcal{S}\rangle$,
         $C=\{\varphi_1,\ldots,\varphi_m\}$
\ENSURE $\Pi'=\langle\mathbf{P}',\mathbf{O},\mathcal{A}',I',G',\mathcal{S}\rangle$
\STATE $\mathcal{A}' \leftarrow \text{copy}(\mathcal{A})$;\quad
       $\mathbf{P}' \leftarrow \mathbf{P}$
\FOR{$j=1,\ldots,m$}
  \STATE $\mathcal{D}_j \leftarrow \textsc{LTL$_f$2DFA}(\varphi_j)$
  \STATE $q_j^{\mathrm{start}} \leftarrow \delta_j(q_j^0,L(I))$
  \STATE add $M_j^k$ to $\mathbf{P}'$ for every $q^k_j\in Q_j$
  \STATE add $\mathsf{Is}\text{-}o$ to $\mathbf{P}'$ for all
         $o\in\mathrm{Obj}(\varphi_j)$
\ENDFOR
\STATE add $\mathit{Valid}$ to $\mathbf{P}'$
\FORALL{$a\in \mathcal{A}'$}
  \FORALL{$j$ and $(q_j^k,g,q_j^\ell)\in\mathcal{T}_j$ with $q_j^k\neq q_j^\ell$}
      \STATE $r \leftarrow \mathrm{Reg}_a(g)$
      \IF{$q_j^\ell=q_j^\bot$}
        \STATE Add $(M_j^k \land r) \rhd (\neg \mathit{Valid})$ to $a$
      \ELSE
        \STATE Add $(M_j^k \land r) \rhd (\neg M_j^k \land M_j^\ell)$ to $a$
      \ENDIF
  \ENDFOR
\ENDFOR
\STATE $I'\leftarrow I\cup \{M_j^{\mathrm{start}}\mid j=1,\ldots,m\} \cup \{\mathsf{Is}\text{-}o(o)\mid o\in\mathrm{Obj}(C)\}$
\IF{$q_j^{\mathrm{start}} \neq q_j^{\bot}$ for every $j$}
    \STATE $I' \leftarrow I' \cup \{\mathit{Valid}\}$
\ENDIF
\STATE $G'\leftarrow G\wedge\bigwedge_j\mathrm{Acc}_j \wedge\mathit{Valid}$
\RETURN $\langle\mathbf{P}',\mathbf{O},\mathcal{A}',I',G',\mathcal{S}\rangle$
\end{algorithmic}
\end{algorithm}
\paragraph{Monitor state and initialization.}
For every $q_j^k \in Q_j$, SAM-TD introduces a fluent $M_j^k$, meaning that monitor $j$ currently occupies state $q_j^k$.
If $q_j^{\mathrm{start}}=q_j^k$, then $M_j^k$ is initially true. 
Exactly one such fluent is initialized for each monitor. 
SAM-TD then adds all required static identity facts $\mathsf{Is}\text{-}o(o)$. 
The shared fluent $\mathit{Valid}$ is initially true if and only if no monitor starts in its sink state.
\paragraph{Action-synchronous monitor updates.}
For each action schema $a$ and each non-self DFA edge $(q_j^k,g_j^{k\ell},q_j^\ell)$
with $q_j^\ell\neq q_j^\bot$, SAM-TD adds the conditional effect
\begin{align*}
 \bigl(M_j^k\land\mathrm{Reg}_a(g_j^{k\ell})\bigr)
 \;\rhd\;
 \bigl(\neg M_j^k\land M_j^\ell\bigr).
\end{align*}
For each transition into a sink state $q_j^\bot$, SAM-TD deletes the $\mathit{Valid}$ token (Token Destruction Mechanism)
\begin{align*}
 \bigl(M_j^k\land\mathrm{Reg}_a(g_j^{k\bot})\bigr)
 \;\rhd\;
 \neg\mathit{Valid}.
\end{align*}
No action adds $\mathit{Valid}$. We augment the planning goal $G$ as follows:
\begin{align*}
 \mathit{Acc}_j
   = \bigvee_{q_j^k\in F_j}M_j^k,
 \qquad
 G'=G\land\mathit{Valid}\land\bigwedge_{j=1}^{m}\mathit{Acc}_j.
\end{align*}
The original streams are copied unchanged. The full SAM-TD compilation is shown in Algorithm \ref{alg:gr-compile}.
\subsection{Validity Monitor and Heuristic Pruning}\label{subsec:VMA}
The fluent $\mathit{Valid}$ is the planning representation of a two-state
monitor $\mathcal{V}_C$ with states $v_{\mathrm{ok}}$ and $v_{\mathrm{bad}}$
\begin{align*}
    \mathcal{V}_C = \big\langle \{v_{\mathrm{ok}},\, v_{\mathrm{bad}}\},\; \delta_{\mathcal{V}},\; v^{\mathrm{start}} \big\rangle,
\end{align*}
as shown in Fig.~\ref{fig:samtd-pipeline} (Token Destruction Mechanism).
$\delta_{\mathcal{V}}$ moves $v_{\mathrm{ok}}$ to $v_{\mathrm{bad}}$ exactly when any $\mathcal{D}_j$ takes a sink transition; $v_{\mathrm{bad}}$ is absorbing, and $v^{\mathrm{start}} = v_\mathrm{ok}$ if and only if no monitor starts in its sink state.
The PDDL fluent $\mathit{Valid}$ denotes $v_{\mathrm{ok}}$. 
Token Destruction exposes constraint violation to heuristics such as $h^{\text{FF}}$. 
Since no action can add $\mathit{Valid}$, under delete-relaxation heuristics this positive goal fluent is unachievable from any state in which it has been deleted, so $h^{\text{FF}}$ assigns such states infinite cost.
In the PDDLStream search architecture, this allows the symbolic branch to be discarded before it is selected for geometric object sampling. 
\section{Experiments}\label{sec:experiments}
\subsection{PDDLStream Experiment Setup}\label{subsec:PDDLStream-Experiment-Setup}
We use three PDDLStream environments: \emph{Kitchen} from \cite{Garrett2020}, \emph{Tabletop Manipulation} derived from \cite{wu2025selp}, and our own \emph{ZoneSort} environment. 
Each environment has a task-specific LTL$_f$ constraint set (see \suppref{app:exp1}). 
We use the PDDLStream \textit{Adaptive} algorithm \cite{Garrett2020} as the solver for the compiled planning problem.
We replay all solver-returned plans to verify executability, achievement of the original goal, and constraint satisfaction.
\paragraph{Experiment~1: Arbitrary LTL$_f$ constraint scaling}
We compile five different orderings of task-specific constraint sets (\suppref{app:exp1}), one constraint at a time.
At any fixed number of constraints $k$, different orderings enforce different subsets of the task-specific constraint set.
\paragraph{Experiment~2: Adversarial Constraint Scaling}
We compile adversarial LTL$_f$ constraints which are designed to reject by an unconstrained plan and force re-planning with each added LTL$_f$ constraint. Details on the adversarial constraint design are in the supplementary material. 
\paragraph{Experiment~3: (Ablation) Compiling without the Token Destruction Mechanism}
A natural variant of SAM-TD instead blocks violations at the precondition: for each regressed sink-transition guard, the negated guard is conjoined into the action's precondition, so that any constraint-violating instantiation becomes inapplicable. 
We call this variant SAM-PB (precondition blocking).
The supplementary formalizes precondition blocking.  
We carry out Experiment~2 with SAM-PB using identical LTL$_f$ formulas, solver, planning budgets, and memory limits and compare its behavior with SAM-TD.
\subsection{Classical PDDL Setup}\label{subsec:setup-pddl}
Since no prior compilation technique supports streams, we compare SAM-TD against existing methods on six published PDDL3 tasks \cite{Gerevinietal2009}, two each from Openstacks, Rovers, and Storage, enforcing the first $k$ trajectory constraints at five different levels per task, and up to $k=128$ on Storage, with a 180-second compile-plus-search budget, a 3\,GiB memory limit, and \texttt{lama-first} heuristic planner.
We evaluate SAM-TD against BM06 \cite{Baier2006}, TB15 \cite{Torres2015}, TCORE \cite{Bonassi2021On}, LiftedTCORE and LCC \cite{Mantenoglou2026}, and Plan4Past \cite{DeGiacomoetal2022}.
\subsection{Results and Discussion}
\label{subsec:results}
\begin{figure}[t]
\centering
\includegraphics[width=\columnwidth]{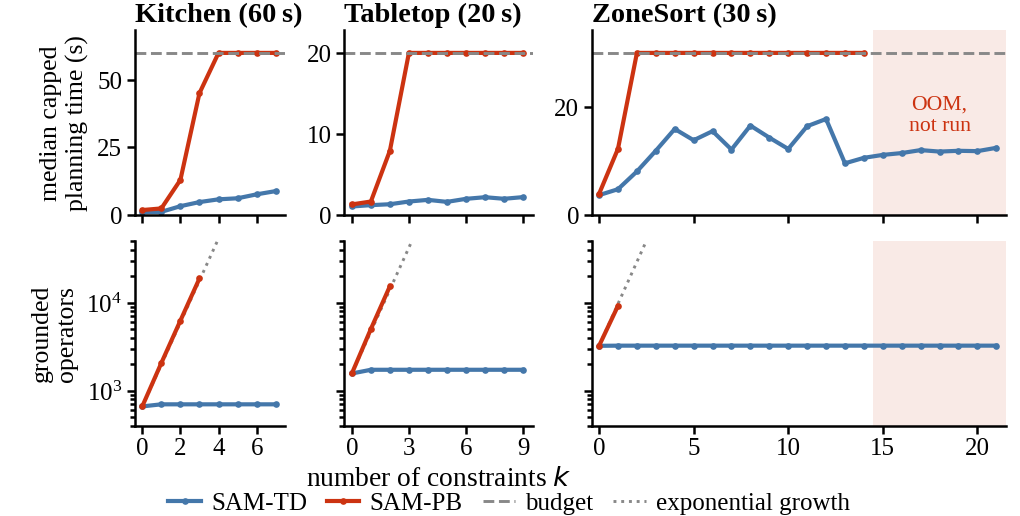}
\caption{Encoding ablation on the Experiment~2 prefixes: median
planning time for SAM-TD and SAM-PB, with timeouts capped at the
per-environment budget (dashed). Capped points are right-censored. The
ZoneSort SAM-PB arm has one accepted repeat; it exhausts memory at
$k{=}15$ before any Fast Downward invocation, and later prefixes were
not run (shaded).}
\label{fig:sampb-ablation}
\end{figure}

\paragraph{Temporal constraint compilation is linear in the number of constraints and action schemas}
We separate the cost of constructing the DFAs from the cost of compiling them into the planning task.
Given the DFAs, SAM-TD introduces at most $|\mathcal{A}| \sum_{j = 1}^{m} |\mathcal{T}_j|$ transition constructs, where $|\mathcal{A}|$ is the number of lifted action schemas and $\mathcal{T}_j$ is the set of transitions of the $j$th DFA.
For a fixed domain and bounded-size DFAs and guards, the compilation is therefore linear in the number of constraints $m$.
The measurements agree (Fig.~\ref{fig:pddlstream-scaling}): median compilation time grows linearly at roughly 30 to 80 milliseconds per added constraint. 
Additionally, Fig.~\ref{fig:pddlstream-scaling}'s runtime breakdown shows that the compilation layer is never the bottleneck with increasing constraints. 
\paragraph{Grounding incurs the highest cost}
Fig.~\ref{fig:pddlstream-scaling} shows the two constraint scaling experiment results. 
For both experiments, the main cost for compilation and planning comes from grounding the action schema into action operators (FD translate). 
The PDDLStream \textit{Adaptive} algorithm alternates between planning over an optimistic abstraction, where placeholder objects are used for geometric objects which have not yet been sampled, and calling geometric samplers to realize the resulting plan skeleton.
When sampling cannot satisfy the skeleton plan's geometric requirements, the failure is recorded and a new skeleton is planned which invokes a new FD translate.
Since every added constraint enlarges the task, the grounding cost incurred during every skeleton plan generation also increases.
The price of search and sampling, however, remains flat since we prune violating branches before expending geometric sampling on them.
\paragraph{Precondition blocking leads to ground action operator explosion}
Fig.~\ref{fig:sampb-ablation} shows the ablation experiment (Experiment 3) between SAM-TD and SAM-PB. 
SAM-PB exhausts the planning budget for small $k$ across Kitchen and Tabletop, and exhausts the pre-allocated memory budget (3 GiB) \textit{before} planning even begins for ZoneSort.
The exponential ground operator blowup for precondition blocking arises because precondition blocking yields a logical conjunction of disjunctions that DNF-normalizing translators such as Fast Downward compile into one operator copy per combination of disjuncts, yielding up to $d^{m}$ ground copies, where $d$ bounds the number of disjuncts in each negated guard. 
This ablation shows the value of the Token Destruction mechanism which allows violations to live in conditional effects that ground linearly and are pruned by $h^{\text{FF}}$
\paragraph{Competitive scaling on published PDDL3 constraints}
Although SAM-TD is designed for stream-based TAMP, its compilation remains competitive when streams are absent. 
Table~\ref{tab:pddl3} compares the seven methods on the 30 task-constraint-level instances. 
SAM-TD, BM06, TCORE, and Plan4Past solve all 30 instances of the planning problem.
On the 11 instances solved by every method, SAM-TD acheives the second lowest geometric-mean total time.
Thus, our results indicate that stream compatibility does not prevent competitive performance on classical PDDL.
\begin{table}[t]
\centering
\scriptsize
\setlength{\tabcolsep}{1.8pt}
\begin{tabular}{@{}lccccc@{}}
\toprule
& & \multicolumn{3}{c}{max constraints $k$} & \\
\cmidrule(lr){3-5}
Compiler & Solved
& \shortstack{Openst.\\(24/29)}
& \shortstack{Rovers\\(30/56)}
& \shortstack{Storage\\(128/128)}
& $t_{\cap}$ (s)\\
\midrule
\textbf{SAM-TD (ours)}
  & \textbf{30/30}
  & \textbf{24/29}
  & \textbf{30/56}
  & \textbf{128/128}
  & \textbf{1.88}\\
BM06        & 30/30 & 24/29 & 30/56 & 128/128 & 2.71\\
TB15        & 15/30 & 4/4   & 4/8   & 4/64    & 9.55\\
TCORE       & 30/30 & 24/29 & 30/56 & 128/128 & 2.16\\
LiftedTCORE & 22/30 & 24/29 & 30/56 & 1/1     & 6.77\\
LCC         & 22/30 & 24/29 & 30/56 & 1/1     & \textbf{1.15}\\
Plan4Past   & 30/30 & 24/29 & 30/56 & 128/128 & 2.47\\
\bottomrule
\end{tabular}
\caption{Published PDDL3 scaling over six tasks and five levels of
constraint prefixes per task. Domain entries report the max verified
$k$ for the two tasks in order: Openstacks p61/p75, Rovers p60/p90, and
Storage p19/p24. $t_{\cap}$ is the geometric-mean compiler-plus-planner
time on the instances solved by all seven methods.
Complete outcomes, pairwise runtimes, and representation sizes are in supplementary material.}
\label{tab:pddl3}
\end{table}
\section{Summary and Future Work}\label{sec:conclusion}
We present Synchronous Action Monitoring with Token Destruction (SAM-TD), the first compilation method for enforcing arbitrary Linear Temporal Logic over finite traces (LTL$_f$) constraints in stream-based robotics Task and Motion Planning.
SAM-TD translates each LTL$_f$ constraint into a Deterministic Finite Automaton (DFA), and encodes the automaton's transitions as conditional effects on the action schemas, synchronously advancing the DFA state inside the compiled task.
Any action that violates a temporal constraint deletes a single shared validity token, so violating branches are pruned by standard delete-relaxation heuristics.
Our experiments provide the first demonstration of LTL$_f$-constrained stream-based TAMP. 
Compilation time grows linearly with the number of constraints (for bounded-size DFAs and guards).
The ablation shows that the Token-Destruction mechanism keeps action grounding tractable.
This work shows that temporal constraints in stream-based TAMP can be enforced by compilation alone, with end-to-end cost governed by the planner's own grounding rather than by search, geometric sampling, or the compilation logic.
For future work, we plan to extend SAM-TD to mobile and multi-robot TAMP, and to bi-level planners that combine reinforcement-learning skills and samplers with heuristic task-level search, which SAM-TD constrains directly.

\clearpage
\bibliography{aaai2027}
\clearpage

\appendix
%
%
%
%

\section{Why Classical Temporal Compilations Are Incompatible with Stream-Based Instantiation}
\label{app:instantiation-incompatibility}

Classical planners can ground eagerly: a translator enumerates the closed, typed object set and emits every ground operator before search begins. 
Stream-based solvers cannot, since the object set grows as streams produce values, so PDDLStream replaces the translator with its own instantiation layer (\S2.4 of the main paper). 
The replacement changes which PDDL constructs a compilation may rely on. 
This section describes those semantics at the level a compilation sees them and traces how each classical compilation artifact falls outside them. 

\paragraph{Fact-driven instantiation.}
Rather than enumerating a universe, the instantiation layer lets facts drive grounding. A binding $\theta$ for a schema $a$ is assembled by matching the static atoms of $\mathrm{pre}(a)$ against the optimistic fact set $I^*$; both initial static facts and stream-certified facts qualify, the latter because a certified fact, once generated, is never deleted. Writing $\mathrm{pre}_{\mathrm{stat}}(a)$ for these atoms and $\mathrm{vars}(\bar t)$ for the parameters occurring in the argument tuple $\bar t$, the layer instantiates $a$ only when every parameter has such a source of candidate objects:
\begin{equation}
 \mathrm{params}(a) \;\subseteq \bigcup_{p(\bar t)\,\in\,\mathrm{pre}_{\mathrm{stat}}(a)} \mathrm{vars}(\bar t).
 \label{eq:coverage}
\end{equation}
A geometric parameter is therefore bound exactly when some certified atom mentions it: $\mathrm{Kin}(b, q, p, g)$ in a precondition supplies candidates for the pose, grasp, and configuration as streams produce them, which is what allows optimistic placeholders for not-yet-sampled objects to participate in grounding at all \cite{Garrett2020}. Fluent preconditions play no part in binding and are checked during search. Conditional-effect conditions likewise pass through instantiation untouched and are evaluated as ordinary state lookups by the search engine. This asymmetry, binding-time preconditions against search-time effect conditions, decides where a compilation may and may not place its monitoring logic.

\paragraph{What falls outside the fragment.}
Three constructs that classical compilations lean on have no place in this discipline, and they are exactly the three named in \S2.4. 
\begin{enumerate}
    \item \textit{R1: Equality check is not supported.}
    An atom $(= x, o)$ is certified by no fact, so it can neither generate nor filter a binding, and the layer provides no equality test.
    \item \textit{R2: Parameter--constant matching cannot be written directly.}
    An atom that names an object, $\mathit{moved}(\mathit{blue})$ rather than $\mathit{moved}(?b)$, binds nothing under Eq.~\eqref{eq:coverage}; empirically, schemas whose preconditions carry such atoms fail to instantiate in the reference implementation.
    \item \textit{R3: No action parameter may be left unbound.}
    A new variable introduced by a compilation, appearing in no static atom of the host schema, violates Eq.~\eqref{eq:coverage} and blocks instantiation of the entire schema.
    \item \textit{R4: Negated sink fluents not accepted by the goal.}
    Goals are conjunctions of positive conditions, so a negated sink fluent cannot be required at the end of the plan.
\end{enumerate}
\paragraph{Where classical compilations collide.}
Each classical artifact meets one of these walls. Compilations in the style of \citet{Baier2006} and \citet{DeGiacomoetal2022} write monitor updates as conditional effects over ground atoms; the atoms themselves would survive instantiation, but the surrounding pipeline does not (\textit{R2}). Relating the named object to the acting parameter instead requires the equality test the fragment lacks, the construct on which LiftedTCORE's regression depends \cite{Mantenoglou2026} (\textit{R1}).
Grounding the task up front, as TCORE does \cite{Bonassi2021On}, presupposes the closed object universe that streams remove. 
Moving monitor updates into dedicated synchronization actions \cite{Torres2015, Camachoetal2017}, or bookkeeping actions as in LCC \cite{Mantenoglou2026}, avoids all three constructs but violates the second requirement of \S2.4 instead: the inserted actions have no stream support, add trace steps that change the semantics of step-sensitive operators, and break the plan--sample--refine loop. 
In our development, every repair of one construct landed on another.

\section{Discussion of SAM-TD's Formal Properties}
\label{app:formal-properties}

The main paper makes two claims that this section discusses in more detail. 
First, \S2.4 states that SAM-TD satisfies the three stream-specific requirements by construction. 
Second, \S3.1 asks for a compilation whose solutions are exactly the solutions of $(\Pi, C)$. Rather than developing formal theorems and proofs, we walk through the constructions of \S3 and Algorithm~1 of the main paper and point out, for each claim, the structural facts of the compilation that it rests on. The regression step specializes the standard correctness argument for action regression \cite{Rintanen2008}, and the monitor bookkeeping follows the usual automata-product pattern of temporal-goal compilations \cite{Baier2006}. What is specific to SAM-TD, and what this discussion therefore focuses on, is how these components interact with the static identity predicates, the shared $\mathit{Valid}$ token, and an object universe that streams keep extending during search.

\paragraph{Scope.}
Throughout, we rely on conditions that are implicit in the main paper. Constraints are stated over ground task-level atoms: each $AP_j$ is finite, every object named in a formula is known when SAM-TD runs, and stream-certified predicates never occur in $AP$, so extending a state with stream-certified facts never changes its trace letter. Note that this does not require stream outputs to be known at compilation time. Poses, grasps, and trajectories may still enter as action arguments after compilation; only the constants written in the formulas themselves must exist beforehand. Original actions are deterministic, all compilation symbols ($M_j^k$, $\mathsf{Is}\text{-}o$, $\mathit{Valid}$) are fresh, and conditional effects follow standard PDDL semantics: conditions are read in the pre-action state and all selected effects apply simultaneously. Finally, each DFA merges all states from which no accepting state is reachable into the single sink $q_j^\bot$. Merging them preserves the accepted language, since none of them can reach an accepting state anyway, and the merged state is absorbing: a transition leaving it would make an accepting state reachable after all.

\paragraph{The compiled monitors track the DFA runs.}
Three observations chain together. First, the identity predicates behave like equality. $\mathsf{Is}\text{-}o$ receives the single initial fact $\mathsf{Is}\text{-}o(o)$ and is then never added or deleted by any action or certified by any stream, so in every reachable state $\mathsf{Is}\text{-}o(x)$ holds exactly when $x = o$. This remains true for objects that streams introduce later, since no stream certifies the fresh predicate. The static atoms therefore implement the parameter--constant test that the instantiation layer forbids us (\textit{R1}, \textit{R2}) from writing as an equality literal. Second, with the identity atoms in place, regression is exact in every reachable compiled state: $s \models \mathrm{Reg}_a(g)[\theta]$ holds exactly when $\gamma(s, a(\theta)) \models g$ for every applicable ground instance $a(\theta)$. The atomic case is the familiar reading of \S3.3, the atom holds after the action exactly when the action adds it or it already held and is not deleted, and compound guards follow because regression distributes over the Boolean connectives \cite{Rintanen2008}. Third, each action selects exactly one outgoing edge per monitor. The guards leaving any DFA state partition the alphabet $2^{AP_j}$ (\S3.2), regression transfers that partition to the pre-action state, and the conjoined $M_j^k$ switches off every edge whose source state the monitor does not occupy. A self-loop emits no effect, a forward edge deletes its source fluent and adds its target, and a sink edge deletes the token.

Putting these together yields the invariant behind the whole method. Write $s'_i$ for the compiled state after the first $i$ domain actions and $\rho_{j,i}$ for the state of $\mathcal{D}_j$ after reading $L(s_0) \cdots L(s_i)$, so that $\rho_{j,0} = q_j^{\mathrm{start}}$. Along every executable prefix,
\begin{align}
 \mathit{Valid} \in s'_i
 \;&\Longleftrightarrow\;
 \rho_{j,i} \neq q_j^\bot \;\text{ for all } j,
 \label{eq:token-tracks-sinks}\\
 \mathit{Valid} \in s'_i
 \;&\Longrightarrow\;
 \big(M_j^k \in s'_i \Leftrightarrow q_j^k = \rho_{j,i}\big) \text{ for all } j, k.
 \label{eq:monitor-tracks-run}
\end{align}
A sink edge deletes only $\mathit{Valid}$, so after a violation the monitor fluent may lag its run. This is harmless: the token is already gone, no action re-adds it, and by absorption the run stays in $q_j^\bot$, so both sides of Eq.~\eqref{eq:token-tracks-sinks} remain false forever.

\paragraph{Exactness of the compilation.}
Correctness now reads directly off the compiled goal $G' = G \wedge \mathit{Valid} \wedge \bigwedge_j \mathrm{Acc}_j$. A plan reaching $G'$ satisfies the original goal, since injected effects touch only fresh fluents, and has kept the token, so by Eq.~\eqref{eq:token-tracks-sinks} no run entered a sink and by Eq.~\eqref{eq:monitor-tracks-run} its monitors are current and sit in accepting states; by DFA correctness \cite{DeGiacomo2013} its trace satisfies every $\varphi_j$. Conversely, a plan whose trace every $\mathcal{D}_j$ accepts can never visit the absorbing, non-accepting sink, so its token survives and its monitors end in $F_j$. The same action sequence therefore executes in $\Pi'$ and reaches $G'$: preconditions and original effects are untouched, and the injected effects cannot conflict because at most one non-self edge fires per monitor. Solutions of $\Pi'$ are therefore exactly the solutions of $(\Pi, C)$, the empty plan included: it is accepted precisely when $I$ satisfies $G$ and every $q_j^{\mathrm{start}}$ is accepting. And because the streams and their sampling assumptions are untouched, any PDDLStream algorithm that is sound and (probabilistically) complete on the compiled fragment keeps that guarantee for $(\Pi, C)$. One subtlety deserves emphasis: Token Destruction only catches irrecoverable prefixes. A liveness obligation such as $\mathsf{F} p$ can remain recoverable through the entire trace and still be unmet when the plan stops. That failure mode never touches the token and is caught, exactly once, by the $\mathrm{Acc}_j$ conjunct in the goal.

\paragraph{Schema-level, stream-compatible encoding.}
Algorithm~1 copies each schema and appends conditional effects; it adds no parameter, no precondition, no action cost, no stream, and no equality literal. Every variable inside an injected condition is already a parameter of the host schema, since it can only enter through one of that schema's effects during regression, so no condition mentions an unbound variable that the instantiation layer would have to resolve. Consequently, a binding grounds the compiled schema exactly when it grounds the original one, and the stream-certified facts a fixed plan needs are identical before and after compilation: SAM-TD neither consumes nor produces stream objects. The compilation also commutes with grounding. Over any finite object universe $\Omega$, grounding the compiled schemas and then evaluating each ground identity atom $\mathsf{Is}\text{-}o(x)$ to its truth value, true exactly for $x = o$, yields the same conditions that ground regression would produce on the grounding of $\Pi$ over $\Omega$. Because this holds for every finite $\Omega$, it holds in particular for the optimistic universe of each PDDLStream iteration \cite{Garrett2020}: when streams extend the universe, the already-compiled schemas ground correctly over the new objects, and the identity tests evaluate correctly on them, to false unless the new object is the named one, which streams never produce. Thus, the encoding is stream-compatible without enumerating an object set.

\paragraph{Preservation of the domain-action skeleton.}
SAM-TD introduces no executable action. The map from original to compiled ground actions is the identity on name and binding, so a solution has the same actions in the same order, with the same length and cost, and the projected task-level trace contributes exactly one letter per domain action after the initial one. Step-sensitive operators such as $\mathsf{X}$ are therefore evaluated at the same trace positions before and after compilation, and the skeleton handed to the refinement phase is the one the domain author wrote: no synchronization actions need to be inserted during search or stripped from returned plans.

\paragraph{Sampling-aware pruning.}
On reachable compiled states, losing the token is synonymous with violation by Eq.~\eqref{eq:token-tracks-sinks}. Such a state is a genuine dead end, since $\mathit{Valid}$ is a goal conjunct and no action re-adds it. It is also a delete-relaxation dead end: relaxation drops delete effects but never introduces facts absent from the state, so $\mathit{Valid}$ appears in no layer of the relaxed planning graph and $h^{\text{FF}}$ returns $\infty$ \cite{Hoffmann2001}. An $h^{\text{FF}}$-guided search may therefore discard every post-violation state without losing a single constrained solution. In the plan--sample--refine loop, this discarding happens inside symbolic task-level search, before it can complete into a skeleton, so the refinement phase is never asked to bind geometric placeholders for a branch already known to violate a constraint. 

\paragraph{Compilation size.}
Our compilation introduces one monitor predicate per DFA state, one identity predicate and initial fact per object in $\mathrm{Obj}(C)$, one $\mathit{Valid}$ fluent, and at most $|\mathcal{A}| \sum_{j=1}^{m} |\mathcal{T}_j|$ injected conditional effects. For a fixed domain and bounded-size DFAs and guards, the compilation is therefore linear in $m$, matching the measurements in \S4.3. This bound only accounts for SAM-TD's compilation and does not take into account a downstream translator's normalization or grounding strategy, which \S4.3 measures separately.

%
%
%
%

\section{SAM-PB: Precondition Blocking}
\label{app:sam-pb}

SAM-PB is the encoding ablation of Experiment~3 (\S4.1 of the main paper). It shares its entire front end with SAM-TD: the same DFAs and pre-evaluated start states (\S3.2), the same identity predicates and lifted regression (\S3.3), and the same forward-edge conditional effects that advance each monitor fluent $M_j^k$ in synchronization with the action (\S3.4). The two compilation methods differ only in sink transition encoding. 
SAM-PB conjoins the negated regressed guard into the host action's precondition. For each action schema $a$ and each sink transition $(q_j^k, g_j^{k\bot}, q_j^\bot) \in \mathcal{T}_j$,
\begin{equation}
 \mathrm{pre}(a) \;\leftarrow\; \mathrm{pre}(a) \,\wedge\, \neg\big(M_j^k \wedge \mathrm{Reg}_a(g_j^{k\bot})\big).
 \label{eq:pb-precondition}
\end{equation}
By the exactness of regression (\S3.3), the added conjunct is false in precisely those states where executing $a$ would drive $\mathcal{D}_j$ into its sink, so every violating instantiation becomes inapplicable, where SAM-TD would leave it applicable but make the goal unreachable.

With sink transitions unreachable, the $\mathit{Valid}$ token has nothing to record and is dropped: SAM-PB sets $I' = I \cup \{M_j^{\mathrm{start}} \mid j = 1, \ldots, m\} \cup \{\mathsf{Is}\text{-}o(o) \mid o \in \mathrm{Obj}(C)\}$ and $G' = G \wedge \bigwedge_{j=1}^{m} \mathrm{Acc}_j$. If some $q_j^{\mathrm{start}} = q_j^\bot$, the compiled problem is unsolvable. 
The two encodings accept exactly the same plans. A solution of either compiled problem never takes a sink transition, in SAM-PB because Eq.~\eqref{eq:pb-precondition} forbids it and in SAM-TD because the deleted $\mathit{Valid}$ token makes the goal unreachable, and the discussion of \S\ref{app:formal-properties} otherwise carries over unchanged. The ablation therefore isolates a purely representational choice: the same regressed guard, placed in a precondition versus a conditional effect.

That placement is what Experiment~3 measures. $\mathrm{Reg}_a(g_j^{k\bot})$ is in general a disjunction of conjunctive cases, so its negation in Eq.~\eqref{eq:pb-precondition} is a conjunction of disjunctions, and the precondition accumulates one such block per sink edge across all $m$ constraints. DNF-normalizing translators such as Fast Downward multiply these blocks out, one operator copy per combination of disjuncts, yielding up to $d^m$ ground copies where $d$ bounds the number of disjuncts in each negated guard. SAM-TD carries the identical guard inside a conditional effect, which grounds once per edge; this difference alone accounts for the operator explosion reported in \S4.3.

%
%
%

\section{Adversarial Constraint Design}
\label{app:adversarial-design}

This appendix details the construction behind Experiment~2 (\S4.1 of the main paper). For each environment, we first solve the unconstrained problem and independently verify the returned plan; this reference plan fixes an ordered sequence of task-level events (placement and manipulation milestones such as $\mathit{placed}(\cdot)$ and $\mathit{moved}(\cdot)$). The adversarial catalog is then enumerated deterministically from that sequence. For every ordered pair of reference events, $e_a$ occurring before $e_b$, we add the ordering reversal $(\neg\, e_a)\ \mathsf{U}\ e_b$, which forbids $e_a$ until $e_b$ has occurred; afterwards, we append every catalogued safety invariant $\mathsf{G}(\neg f)$ whose atom $f$ the reference trace makes true. Each retained formula is independently checked, by running its DFA on the reference trace, to reject the common reference plan, and a single independently verified witness plan that satisfies the full catalog proves that every nested prefix remains feasible. The scaling runs then enforce the literal prefixes $\{\varphi_1, \ldots, \varphi_k\}$ for $k = 0, \ldots, K_{\max}$, with three repeats per depth. Adversarial thus has a precise, limited meaning in our experiments: every formula rejects the common unconstrained reference plan, so the unconstrained solution never survives compilation; it need not reject the constrained plan found at depth $k-1$.

%
%
%
%

\section{Classical PDDL3 Scaling Curves}
\label{app:pddl3}

\begin{figure*}[t]
\centering
\includegraphics[width=\textwidth]{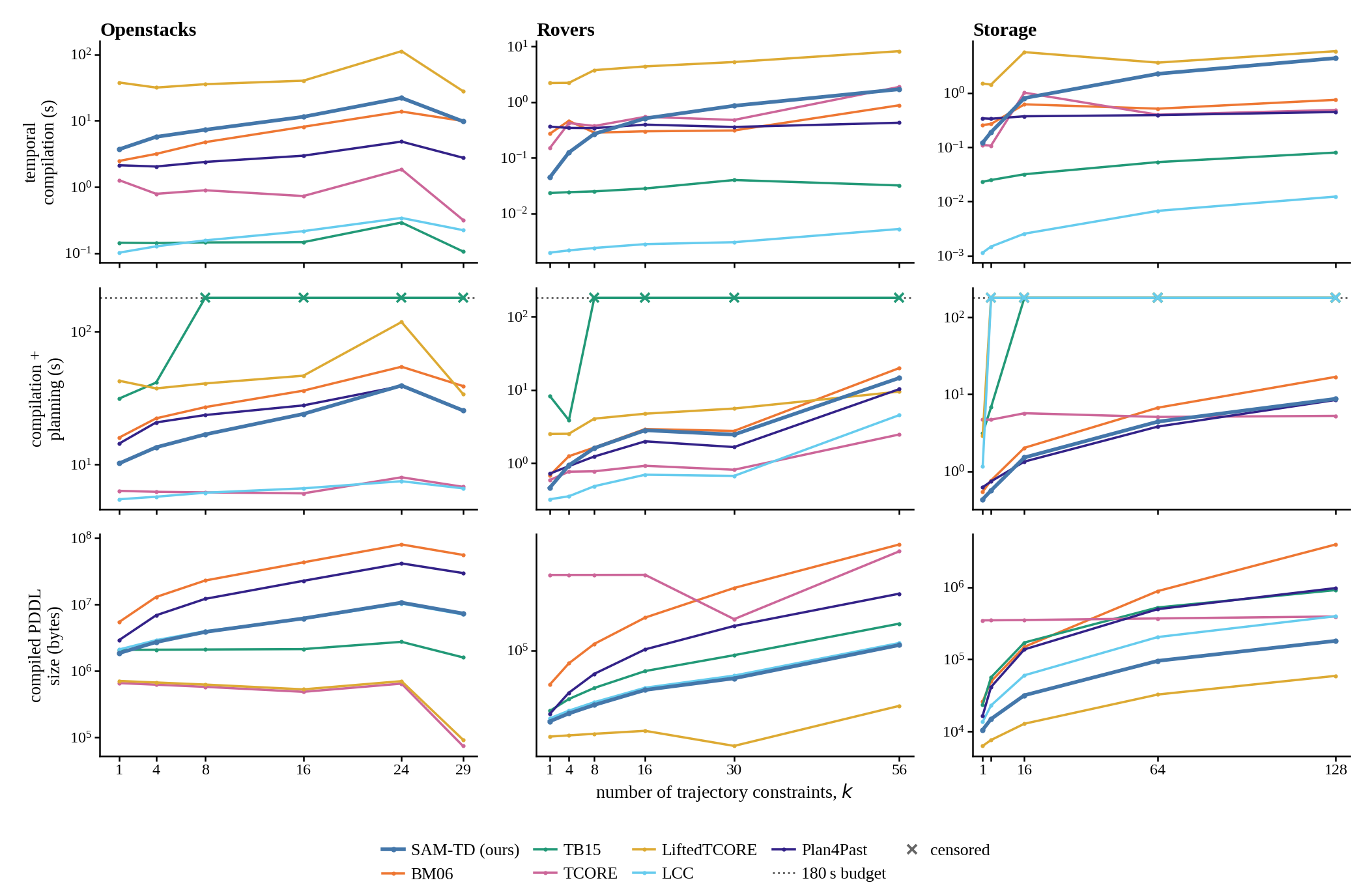}
\caption{Published PDDL3 trajectory-constraint scaling on the six classical tasks of \S4.2 (two per domain), with medians taken over the tasks available at each exact $k$. Top row: temporal compilation time. Bottom row: total compile-plus-search time, on logarithmic axes. After a method's first failure, its curve is held at the 180-second budget and marked with crosses (right-censored); those flat segments are not measured runtimes.}
\label{fig:pddl3-scaling}
\end{figure*}

Fig.~\ref{fig:pddl3-scaling} complements Table~1 of the main paper with the full scaling curves behind its coverage counts. The bottom row shows that the aggregate outcomes are structured. SAM-TD, BM06, TCORE, and Plan4Past complete all 30 task-constraint cells with totals that grow gently in $k$; TB15 is right-censored earliest in all three domains; and LiftedTCORE and LCC finish Openstacks and Rovers but exhaust the 3\,GiB memory budget immediately beyond Storage's first level, which is exactly the $1/1$ entry Table~1 reports for both. The top row separates the temporal compilation share: for SAM-TD it stays well below total time at every depth, consistent with the finding of \S4.3 that grounding rather than compilation dominates end-to-end cost, while LiftedTCORE shows the steepest compiler-side growth, exceeding one hundred seconds at the deepest Openstacks level.

Across the 210 method rows of the seven-method matrix, 179 solved and passed independent verification, 6 timed out, and 25 exhausted memory. Every returned plan was replayed against the original task, the original goal, and the exact source LTL$_f$ formulas, with compiled automaton facts never used as a correctness oracle. Because the protocol runs one timing measurement per exact task-and-$k$ cell, and the two tasks of a domain end at different prefix lengths, terminal points can aggregate a single task; the curves therefore support descriptive comparison rather than statistical inference.

%
%
%
%
%

\section{PDDLStream Experiment 1: Arbitrary Task-Specific LTL$_f$ Constraint Set}
\label{app:exp1}

For each environment, Experiment~1 (\S4.1 of the main paper) uses one fixed, task-specific LTL$_f$ catalog. Table~\ref{tab:e1-catalogs} lists all three catalogs. The five seed values in Table~\ref{tab:e1-orderings} label predetermined orderings of the same catalog: a run at level $k$ jointly enforces the first $k$ constraints of the selected ordering, so at any fixed $k$ different orderings enforce different subsets of the catalog. Labels $G_i$, $U_i$, and $F_1$ denote safety, strong-until precedence, and eventual-completion constraints, respectively.

\begingroup
\renewcommand{\arraystretch}{1.2}
\begin{table*}
\centering
\footnotesize
\caption{Experiment 1 constraint catalogs for all three environments ($G$: safety, $U$: precedence, $F$: completion). Tabletop abbreviates \texttt{white\_block}, \texttt{red\_block}, \texttt{blue\_block}, and \texttt{green\_block} by $w,r,b,g$; Kitchen abbreviates the cup, spoon, cream cup, sugar cup, stirrer, and block by $c,s,c_r,c_s,t,b$. The six ZoneSort safety formulas cover $b_1,\ldots,b_6$, and its nine precedence formulas form the complete bipartite graph $\{b_1,b_2,b_3\}\rightarrow\{b_4,b_5,b_6\}$.}
\label{tab:e1-catalogs}
\begin{tabular}{@{}l l l p{0.34\textwidth}@{}}
\toprule
Environment & ID & LTL$_f$ formula & Interpretation \\
\midrule
Tabletop
 & $G_1$ & $\mathsf{G}\bigl(\neg\mathit{InDanger}(w)\bigr)$ & White block always avoids the encoded danger route. \\
 & $G_2$ & $\mathsf{G}\bigl(\neg\mathit{InDanger}(r)\bigr)$ & Red block always avoids the encoded danger route. \\
 & $G_3$ & $\mathsf{G}\bigl(\neg\mathit{InDanger}(b)\bigr)$ & Blue block always avoids the encoded danger route. \\
 & $G_4$ & $\mathsf{G}\bigl(\neg\mathit{InDanger}(g)\bigr)$ & Green block always avoids the encoded danger route. \\
 & $U_1$ & $\neg\mathit{Moved}(r)\ \mathsf{U}\ \mathit{Placed}(w)$ & Place white before moving red. \\
 & $U_2$ & $\neg\mathit{Moved}(b)\ \mathsf{U}\ \mathit{Placed}(w)$ & Place white before moving blue. \\
 & $U_3$ & $\neg\mathit{Moved}(g)\ \mathsf{U}\ \mathit{Placed}(r)$ & Place red before moving green. \\
 & $U_4$ & $\neg\mathit{Moved}(g)\ \mathsf{U}\ \mathit{Placed}(b)$ & Place blue before moving green. \\
 & $F_1$ & $\mathsf{F}\bigl(\mathit{Placed}(w)\land\mathit{Placed}(r)\land\mathit{Placed}(b)\land\mathit{Placed}(g)\bigr)$ & Eventually place all four target blocks. \\
\midrule
Kitchen
 & $G_1$ & $\mathsf{G}\bigl(\neg\mathit{CupLifted}(c_s)\bigr)$ & Transfer sugar with the spoon without lifting its source cup. \\
 & $G_2$ & $\mathsf{G}\bigl(\neg\mathit{CupLifted}(t)\bigr)$ & Use the spoon for stirring instead of lifting the stirrer. \\
 & $U_1$ & $\neg\mathit{ScoopDone}(s)\ \mathsf{U}\ \mathit{FillDone}(c)$ & Fill the cup before scooping sugar. \\
 & $U_2$ & $\neg\mathit{PourDone}(c_r,c)\ \mathsf{U}\ \mathit{FillDone}(c)$ & Fill the cup before pouring cream. \\
 & $U_3$ & $\neg\mathit{DumpDone}(s,c)\ \mathsf{U}\ \mathit{PourDone}(c_r,c)$ & Pour cream before dumping the spoon into the cup. \\
 & $U_4$ & $\neg\mathit{CupPlaced}(c)\ \mathsf{U}\ \mathit{FillDone}(c)$ & Fill the cup before placing it. \\
 & $F_1$ & $\begin{aligned}[t]
    \mathsf{F}\bigl(&\mathit{FillDone}(c)\land\mathit{PourDone}(c_r,c)\land\mathit{ScoopDone}(s)\\[-1pt]
    &{}\land\mathit{DumpDone}(s,c)\land\mathit{StirDone}(c)\land\mathit{PushDone}(b)\\[-1pt]
    &{}\land\mathit{CupPlaced}(c)\bigr)
  \end{aligned}$ & Eventually achieve every tracked Kitchen milestone. \\
\midrule
ZoneSort
 & $G_1$ & $\mathsf{G}\bigl(\neg\mathit{InDanger}(b_1)\bigr)$ & Block $b_1$ uses the certified route. \\
 & $G_2$ & $\mathsf{G}\bigl(\neg\mathit{InDanger}(b_2)\bigr)$ & Block $b_2$ uses the certified route. \\
 & $G_3$ & $\mathsf{G}\bigl(\neg\mathit{InDanger}(b_3)\bigr)$ & Block $b_3$ uses the certified route. \\
 & $G_4$ & $\mathsf{G}\bigl(\neg\mathit{InDanger}(b_4)\bigr)$ & Block $b_4$ uses the certified route. \\
 & $G_5$ & $\mathsf{G}\bigl(\neg\mathit{InDanger}(b_5)\bigr)$ & Block $b_5$ uses the certified route. \\
 & $G_6$ & $\mathsf{G}\bigl(\neg\mathit{InDanger}(b_6)\bigr)$ & Block $b_6$ uses the certified route. \\
 & $U_1$ & $\neg\mathit{moved}(b_4)\ \mathsf{U}\ \mathit{sorted}(b_1)$ & Sort $b_1$ before moving $b_4$. \\
 & $U_2$ & $\neg\mathit{moved}(b_5)\ \mathsf{U}\ \mathit{sorted}(b_1)$ & Sort $b_1$ before moving $b_5$. \\
 & $U_3$ & $\neg\mathit{moved}(b_6)\ \mathsf{U}\ \mathit{sorted}(b_1)$ & Sort $b_1$ before moving $b_6$. \\
 & $U_4$ & $\neg\mathit{moved}(b_4)\ \mathsf{U}\ \mathit{sorted}(b_2)$ & Sort $b_2$ before moving $b_4$. \\
 & $U_5$ & $\neg\mathit{moved}(b_5)\ \mathsf{U}\ \mathit{sorted}(b_2)$ & Sort $b_2$ before moving $b_5$. \\
 & $U_6$ & $\neg\mathit{moved}(b_6)\ \mathsf{U}\ \mathit{sorted}(b_2)$ & Sort $b_2$ before moving $b_6$. \\
 & $U_7$ & $\neg\mathit{moved}(b_4)\ \mathsf{U}\ \mathit{sorted}(b_3)$ & Sort $b_3$ before moving $b_4$. \\
 & $U_8$ & $\neg\mathit{moved}(b_5)\ \mathsf{U}\ \mathit{sorted}(b_3)$ & Sort $b_3$ before moving $b_5$. \\
 & $U_9$ & $\neg\mathit{moved}(b_6)\ \mathsf{U}\ \mathit{sorted}(b_3)$ & Sort $b_3$ before moving $b_6$. \\
 & $F_1$ & $\mathsf{F}\bigl(\bigwedge_{i=1}^{6}\mathit{sorted}(b_i)\bigr)$ & Eventually sort all six blocks. \\
\bottomrule
\end{tabular}
\end{table*}
\endgroup

\begingroup
\renewcommand{\arraystretch}{1.15}
\begin{table*}
\centering
\footnotesize
\caption{Five frozen prefix orders per environment. Entries are read from left to right: a run at level $k$ jointly enforces the first $k$ constraints. The seed values are stable pool labels, not run-time shuffle instructions.}
\label{tab:e1-orderings}
\begin{tabular}{@{}l c l@{}}
\toprule
Environment & Seed & Ordered constraints \\
\midrule
Tabletop
 & 11 & $F_1$, $G_1$, $U_4$, $G_4$, $U_1$, $G_3$, $U_3$, $G_2$, $U_2$ \\
 & 17 & $G_2$, $U_4$, $F_1$, $G_4$, $U_3$, $G_3$, $U_2$, $G_1$, $U_1$ \\
 & 23 & $G_2$, $U_1$, $G_4$, $U_3$, $F_1$, $G_1$, $U_2$, $G_3$, $U_4$ \\
 & 31 & $G_4$, $U_4$, $G_3$, $U_2$, $G_1$, $U_3$, $F_1$, $G_2$, $U_1$ \\
 & 47 & $G_4$, $U_4$, $G_1$, $U_2$, $G_2$, $U_1$, $G_3$, $U_3$, $F_1$ \\
\midrule
Kitchen
 & 11 & $F_1$, $U_1$, $U_2$, $G_1$, $U_3$, $U_4$, $G_2$ \\
 & 17 & $G_2$, $U_1$, $U_4$, $U_3$, $G_1$, $U_2$, $F_1$ \\
 & 23 & $U_4$, $G_1$, $F_1$, $U_3$, $U_1$, $G_2$, $U_2$ \\
 & 31 & $U_3$, $G_2$, $U_1$, $F_1$, $U_4$, $G_1$, $U_2$ \\
 & 47 & $U_2$, $U_4$, $G_2$, $G_1$, $F_1$, $U_1$, $U_3$ \\
\midrule
ZoneSort
 & 11 & $F_1$, $U_6$, $U_8$, $G_5$, $U_2$, $U_4$, $G_3$, $U_1$, $G_1$, $U_9$, $G_6$, $U_7$, $G_2$, $G_4$, $U_3$, $U_5$ \\
 & 17 & $U_5$, $U_8$, $G_2$, $F_1$, $U_1$, $G_6$, $U_9$, $U_3$, $G_3$, $U_7$, $U_2$, $G_4$, $G_1$, $U_6$, $U_4$, $G_5$ \\
 & 23 & $U_2$, $U_6$, $G_1$, $G_2$, $U_4$, $U_1$, $F_1$, $U_3$, $G_5$, $U_7$, $U_9$, $G_6$, $G_4$, $G_3$, $U_5$, $U_8$ \\
 & 31 & $U_8$, $G_6$, $U_1$, $U_9$, $U_6$, $G_3$, $U_4$, $G_4$, $G_2$, $F_1$, $U_5$, $U_7$, $G_1$, $G_5$, $U_2$, $U_3$ \\
 & 47 & $G_4$, $U_1$, $U_4$, $U_8$, $U_6$, $G_5$, $U_5$, $G_1$, $U_2$, $G_3$, $U_3$, $G_6$, $F_1$, $U_7$, $G_2$, $U_9$ \\
\bottomrule
\end{tabular}
\end{table*}
\endgroup

\end{document}